\documentclass[a4paper]{cta-author}
\usepackage{verbatim} 
\usepackage{amsmath, graphicx}
\usepackage{mathrsfs}
\usepackage{amsthm} 
\usepackage{epsfig, psfrag} 
\usepackage{amsfonts, amssymb} 

\newcommand{\bS}{\boldsymbol{S}}
\newcommand{\bDS}{\boldsymbol{DS}}
\newcommand{\bT}{\boldsymbol{T}}
\newcommand{\bPF}{\boldsymbol{PF}}
\newcommand{\bNS}{\boldsymbol{NS}}
\newcommand{\bDR}{\boldsymbol{DR}}
\newcommand{\bND}{\boldsymbol{ND}}
\newcommand{\bD}{\boldsymbol{D}}
\newcommand{\bx}{\boldsymbol{x}}
\newcommand{\bp}{\boldsymbol{p}}
\newcommand{\bH}{\boldsymbol{H}}

{}
{}
{}

\DeclareMathAlphabet{\mathbbmsl}{U}{bbm}{m}{sl} 

\begin{document}


\title{Multifunction Cognitive Radar Task Scheduling Using Monte Carlo Tree Search and Policy Networks}

\author{\au{Mahdi Shaghaghi$^{1}$}, \au{Raviraj S. Adve$^{1\corr}$}, \au{Zhen Ding$^{2}$}}

\address{\add{1}{Department of Electrical and Computer Engineering, University of Toronto, Toronto, Canada}
\add{2}{Defence Research and Development Canada (DRDC), Ottawa, Canada}
\email{rsadve@ece.utoronto.ca}}

\begin{abstract}
A modern radar may be designed to perform multiple functions, such as surveillance, tracking, and fire control. Each function requires the radar to execute a number of transmit-receive tasks. A radar resource management (RRM) module makes decisions on parameter selection, prioritization, and scheduling of such tasks. RRM becomes especially challenging in overload situations, where some tasks may need to be delayed or even dropped. In general, task scheduling is an NP-hard problem. In this work, we develop the branch-and-bound (B\&B) method which obtains the optimal solution but at exponential computational complexity. On the other hand, heuristic methods have low complexity but provide relatively poor performance. We resort to machine learning-based techniques to address this issue; specifically we propose an approximate algorithm based on the Monte Carlo tree search method. Along with using bound and dominance rules to eliminate nodes from the search tree, we use a policy network to help to reduce the width of the search. Such a network can be trained using solutions obtained by running the B\&B method offline on problems with feasible complexity. We show that the proposed method provides near-optimal performance, but with computational complexity orders of magnitude smaller than the B\&B algorithm.
\end{abstract}

\maketitle

\section{Introduction}\label{sec:intro}

Recent advances in computing has brought up the opportunity to introduce a new generation of radar systems which incorporate intelligence and cognition into their operation. The term \emph{cognitive radar} was described in~\cite{Haykin06:Cognitive_radar} as a radar system which comprises intelligent signal processing, feedback from the receiver to the transmitter, and also information preservation. This definition was further refined in~\cite{Haykin12:Cognitive} such that along with these three ingredients, a cognitive radar system also prioritizes the allocation of available resources in accordance with their importance. In~\cite[Chapter~5]{Charlish17:CognitiveBook}, a cognitive radar is defined as ``a radar system that acquires knowledge and understanding of its operating environment through online estimation, reasoning and learning or from databases comprising context information, and exploits this acquired knowledge and understanding to enhance information extraction, data processing and radar management''. Our work here is best described by this final definition

A cognitive radar system can be modeled using a layered architecture comprising several information abstraction levels~\cite{Charlish17:CognitiveBook}. Such a modeling is an extension of the JDL model~\cite{White88} and its revised versions~\cite{Steinberg99, Llinas04}. We consider the model introduced in~\cite{Smits08} comprising the physical, signal, object, situation, and impact levels. At the physical layer, antennas and sensors interact with the environment. At the signal level, transmit and receive signals are generated and processed, respectively. The object level deals with the estimation of target properties (location, velocity, etc.) and the scheduling of transmit signals. The estimation of the relations between objects and the allocation of resources occur at the situation level. Finally, at the highest level, the impact of the objects is predicted and used in order to plan actions to achieve mission goals.

We consider radar resource management (RRM) for a cognitive multichannel multifunction radar (MFR). Specifically, we focus on the scheduling problem at the object level. In a cognitive setting, the system learns through interactions with the environment and exploits the acquired knowledge in future decision making. We show how machine learning methods can be used to develop a scheduling module which learns from previous problem instances. 

A MFR handles various functions, such as surveillance, tracking of multiple targets, etc., through the execution of a number of tasks. Each task comprises transmission, waiting, and reception intervals. RRM involves prioritization, parameter selection, and scheduling of such tasks such that the available radar resources, specifically time, frequency, and energy, are assigned to different tasks in an efficient manner~\cite{Jack15:RRM}. RRM becomes especially challenging in overload situations where execution of the radar functions requires resources exceeding the capabilities of the MFR. This may lead to some tasks being delayed or even dropped. 


In a cognitive radar setting, priority assignment and parameter selection for the tasks are dealt with at the situation level. Then, at the object level, the tasks are scheduled on the radar timeline. Various techniques have been developed in the literature for effective resource management for multifunction radars (see for example~\cite{Charlish15:RRM_MFR, Woodbridge07:Sch_MFR, Ponsford01:HFSWR} and references therein). 

The priorities and task parameters such as the dwell time and revisit interval can be determined using rule-based methods for each task individually~\cite{Noyes98:Par_sel} or by joint optimization, across all tasks, of an overall utility function, while accounting for resource constraints~\cite{Charlish15:RRM_MFR,Sinha06:Tracking}. Although such an optimization considers available resources, it does not determine the exact time and the order of execution of the tasks. Effective scheduling is then required to accommodate as many tasks as possible on the radar timeline without dropping tasks or imposing significant delays.

Scheduling can be performed using queue or frame based schedulers~\cite[Chapter~3]{Charlish17:CognitiveBook_v1}. Queue based methods select the next best task from an ordered list of the tasks that are eligible to be executed. For example, such a list can be obtained by sorting the tasks based on their start time or their deadlines, which results in the earliest start time (EST) and earliest deadline (ED) first schedulers, respectively.

Frame based schedulers arrange a set of tasks on a time interval. While the current frame is being executed, the next frame is being calculated. Often heuristics are used, since, in general, optimal task scheduling is an NP-hard problem with attendant exponential computational complexity. For example~\cite{Orman96:framebased}, presents a variety of heuristics, based on the allowed delay for the tasks.

On a separate note, multichannel, e.g., multi-frequency, radars are becoming increasingly viable. Such radars bring the possibility of executing multiple tasks simultaneously on different channels (timelines)~\cite{MultiTask}. However, this additional capability also complicates the already NP-hard problem of task scheduling on a single timeline since tasks must now be assigned to channels as well. In this paper, we consider the problem of task scheduling for a multichannel radar, i.e., one that is able to concurrently execute multiple tasks.

In considering RRM for multichannel radars, we developed heuristic methods as well as the optimal branch-and-bound (B\&B) technique~\cite{MSH17:BnB}. While the B\&B approach is computationally efficient (e.g., compared to exhaustive search), its overall complexity in the general case remains exponential. For anything beyond a relatively small number of tasks and channels, the B\&B algorithm cannot be executed in a reasonable time frame. On the other hand, computationally efficient heuristic methods suffer from relatively poor performance. 

Our goal in this paper is to develop a technique that approaches the performance benefits of the B\&B method and the computational benefits of heuristics. In this regard, we introduce using machine learning (ML) techniques to the RRM problem in a multichannel MFR. Our ML methods \emph{learn} from the previous executions of the B\&B algorithm (in a training phase). The acquired knowledge is then \emph{exploited} in future scheduling problems. 

Specifically, we propose to use the Monte Carlo Tree Search (MCTS) approach, coupled with a neural network, popularized by the AlphaGo and AlphaZero programs~\cite{Silver16:AlphaGo, Silver17:AlphaGoZero}.  The MCTS method, searches for the solution on a tree structure by making sequential decisions towards the most promising direction. In our method, we take benefit from the B\&B dominance rules to eliminate as many nodes as possible from the search tree. At every node of the tree, a policy network is also used to provide a \emph{probability distribution} over the possible choices (considering the probability of leading to the best solution). While such a network is trained using reinforcement learning (e.g.,~\cite{Silver17:AlphaGoZero}), we use supervised learning based on optimal solutions found offline using the B\&B method. We show that our proposed method can obtain near-optimal performance, while the search complexity is reduced by orders of magnitude. 

This paper builds on our recent work in~\cite{MSH17:BnB, MSH18:RadarConf}. As mentioned, in~\cite{MSH17:BnB} we developed the B\&B method for RRM (Section~\ref{sec:BnB_Policy} reviews and extends this work). In~\cite{MSH18:RadarConf}, we used the B\&B approach to train a neural network, an approach we take here as well. However, the approach presented here is completely different. Importantly, the approach in~\cite{MSH18:RadarConf} requires a fixed number of tasks. We expand on these differences in the relevant section of the paper (Section~\ref{sec:BnB_Policy}).

This paper is organized as follows. Section~\ref{sec:ProbFormul} develops the problem at hand. We review the heuristic methods in Section~\ref{sec:Heuristic}. In Section~\ref{sec:BnB} we develop the B\&B algorithm, along with how to generate labeled data for use in the MCTS. The MCTS method is reviewed in Section~\ref{sec:MCTS}. Next, in Section~\ref{sec:BnB_Policy}, we present our proposed method which is based on the MCTS technique combined with the B\&B method and policy networks. The performance and complexity of the aforementioned methods are investigated using numerical simulations in Section~\ref{sec:sim}. Finally, Section~\ref{sec:conclude} concludes the paper.

\section{Problem Formulation} \label{sec:ProbFormul}

The problem at hand is as follows~\cite{MSH18:RadarConf}: consider a multifunction radar with $K$ identical channels. Each channel is associated with a timeline on which tasks can be scheduled. There are $N$ tasks, indexed as $1, \dots, N$, which need to be executed to accomplish the radar missions. Tasks can run concurrently on different channels, but cannot overlap on a given timeline. Furthermore, we consider a non-preemptive scheduling scenario, i.e., once a task starts execution, it cannot be stopped until completion. Each task is associated with a \emph{start time} $r_n, n = 1, \dots, N$ after which the task is ready to be executed and a \emph{deadline} $d_n, n = 1, \dots, N$ after which the task is dropped (with an associated \emph{dropping cost} $D_n$). If executed, each task has a task length (dwell time) of $\ell_n, n = 1, \dots, N$. These parameters are assumed known ahead of the scheduling step.

As an example, consider a tracking task. The start time of the task is determined by the required tracking accuracy and the time when the last measurement was made. The deadline depends on the beamwidth of the radar and the estimated trajectory of the target. We do not perform the task after the target is assumed to have moved out of the radar beam, and the task is therefore dropped with its associated dropping cost. Consequently, further actions may be required to compensate for the dropping.

A scheduled, but delayed, task suffers a \emph{tardiness cost} which is, here, modeled as linearly proportional to the delay; let $e_n$ be the time when task $n$ begins execution; the tardiness cost is given by $w_n (e_n - r_n)$, where $w_n$ is the weight which scales the delay. Let the binary variable $x_{nk}$ indicate if task $n$ is scheduled ($=1$) on the $k-$th timeline or not. Task $n$ is dropped if $x_{nk}=0 \forall k$. Then, the cost associated with the $n$-th task is given by $\sum_{k=1}^K x_{nk} w_n (e_n - r_n) + (1 - \sum_{k=1}^K x_{nk})D_n$. Our joint task selection and scheduling problem is a minimization of the total cost $C$, given by
\begin{equation}
C = \sum_{n = 1}^N \sum_{k=1}^K x_{nk} w_n (e_n - r_n) + (\frac{1}{K} - x_{nk})D_n. \label{eq:objective}
\end{equation}
Here, the optimization variables are $x_{nk}$ and $e_n$. If a task is scheduled, i.e., $x_{nk} = 1$ for some $k$, the execution time $e_n$ needs to be determined as well. Our optimization problem is, therefore,
\begin{eqnarray}
	\{x_{nk}^*, e_n^*\} \hspace{-2mm} &= \underset{x_{nk},e_n}{\text{min}} \sum_{n = 1}^N  \sum_{k=1}^K x_{nk} w_n (e_n - r_n) \nonumber \\
	&  \hspace{30mm} + (1/K - x_{nk})D_n \nonumber \\
	&  \hspace{-20mm} \mathrm{s.t.~} x_{nk} \in \{0, 1\} \nonumber \\
	&  \hspace{-10mm} \sum_{k = 1}^K x_{nk} \le 1 \nonumber \\
	&  \hspace{-10mm} r_n \leq e_n \leq d_n \nonumber \\
	&  \hspace{-10mm} n = 1, \dots, N  \nonumber \\ 
	&  \hspace{-10mm} k = 1, \dots, K \nonumber \\
	&  \hspace{8mm} \text{and no tasks overlap in time}. \label{eq:optimization}  
\end{eqnarray}
The second constraint, $\sum_{k=1}^K x_{nk} \le 1$, ensures that, if scheduled, the task is assigned to a single timeline only.

The scheduling problem without deadlines (and therefore without dropped tasks) is NP-hard~\cite{Lenstra77:Sch_complexity}. It can be shown that the joint task selection and scheduling as given in~\eqref{eq:optimization} is also NP-hard.

\section{Heuristic Methods} \label{sec:Heuristic}

Given the computational complexity of the problem in~\eqref{eq:optimization}, researchers have generally developed suboptimal solutions using heuristic methods~\cite{Yano91:EST_EDD}. One approach is to split the problem into task down-selection and scheduling, and then iterate between these two steps. The criterion we use to choose the tasks in the down-selection phase is the dropping costs which represent the priority of the tasks. Let $\bS$ be the sequence of tasks sorted in non-increasing order based on their dropping costs. Furthermore, let $\bDS$ be the set of down-selected tasks. Initially, $\bDS$ is the empty set. In the down-selection step, we take the task with highest dropping cost from the sequence $\bS$ which has not been selected in the previous iterations, and we add it to the set $\bDS$. 

In the scheduling step, a heuristic method is used to schedule the tasks in $\bDS$ on the $K$ timelines. The schedule is said to be \emph{viable} if the execution times of all the scheduled tasks are before their corresponding deadlines. If the schedule is viable, we keep the last task added to $\bDS$; otherwise, it is removed from $\bDS$. Then, the next iteration is performed by going back to the down-selection step to add a new task from $\bS$ to $\bDS$. This procedure continues until all tasks in $\bS$ are checked (see Table~\ref{alg:heur}).

Given a down-selected set $\bDS$, there are different methods that can be used to schedule the tasks on the timelines. In this paper, we consider two well-known methods: the earliest start time (EST) first and the earliest deadline (ED) first~\cite{Yano91:EST_EDD}. Let $\bT$ be a sequence obtained by sorting the tasks in $\bDS$. In the EST method, tasks are sorted in non-decreasing order based on their start times. In the ED method, tasks are sorted based on their deadlines. Essentially, the EST method schedules tasks as soon as they are ready to be executed, and therefore, reduces delay times. On the other hand, the ED method favors tasks that have earlier deadlines, to reduce the dropping of tasks. 

After obtaining the sequence $\bT$ either using the EST or ED methods, the tasks are sequentially placed on the timelines at the earliest time possible. Let $g_i$ ($1 \leq i \leq K$) be an indicator of the time after which the $i$-th channel is available. For scheduling each task, we select the $k$-th channel such that $k = \arg \min_i (g_i)$. If two or more channels have the same time indicator, the one with the smallest index is chosen. The execution time of the task is set as the maximum of the start time of the task and the time indicator of the channel, i.e., $e_n = \max (r_n, g_k)$. We refer to this procedure as \textit{sequence to schedule mapping}.

After all the tasks from the sequence $\bT$ are scheduled, we can check whether the scheduling is viable or not by checking if all the execution times of the tasks in $\bT$ precede their corresponding deadlines. If the scheduling is not viable, the last down-selected task is removed from $\bDS$ (otherwise, it is kept in $\bDS$). Then, we move forward to the next iteration of task down-selection. 

The description of the EST and ED approaches, so far, follows that in the literature~\cite{Yano91:EST_EDD}. However, the performance of these heuristic methods can be improved by modifying the scheduling step. In the case that the scheduling of tasks in $\bT$ is not viable, we can change the order of tasks in $\bT$ with the hope that such shuffling will result in a viable schedule. Clearly, this can be tried again and again till an optimal solution is found, however with computational complexity. Here, we balance the performance and complexity, we only consider swapping the order of two adjacent tasks until a viable schedule is found or we reach the end of $\bT$.

Specifically, let $a_1, a_2, \ldots, a_{N_T}$ be the sequence of tasks in $\bT$. For $j = 1, 2, \cdots, N_T - 1 $, we swap the order of two consecutive tasks $j$ and $j+1$ in $\bT$ to get the modified sequence $\bT_j = a_1, a_2, \ldots, a_{j+1}, a_j, \ldots$. Then, we check the viability of $\bT_j$. The search stops as soon as a viable schedule is found, in which case, the last down-selected task is kept in $\bDS$, and the viable schedule found is recorded. Then, we go to the next iteration. On the other hand, if we get to the end of $\bT$ with no task swapping resulting in a viable schedule, the last down-selected task is removed from $\bDS$ before going to the next iteration. Note that in the next iteration, after down-selecting a new task, sequence $\bT$ is formed again and task swapping is performed on the new sequence. In other words, the task swapping performed in the previous iteration is ignored.

As we will see, allowing for task swapping significantly improves performance, but at an attendant cost in execution complexity. However, even with task swapping, EST and ED are heuristics without performance guarantees. This motivates the development of the optimal solution - the branch-and-bound technique, described next.

\begin{table}[t]
	\small	
	\vspace{2.5mm}
	\caption{Down-selection and scheduling heuristic}\label{alg:heur}
	\vspace{-2mm}
	\begin{center}
		\begin{tabular}{l}
			\hline \\
			\vspace{-5mm} \\		
			\textbf{Initialization} \\
			\hspace{3mm} Let $\bS$ be the sequence of tasks sorted based on the dropping\\
			\hspace{3mm} costs. \\
			\hspace{3mm} $\bDS \leftarrow \{\}$ \\
			\hspace{3mm} $i \leftarrow 1$ \\
			\hline \\
			\vspace{-5mm} \\		
			\textbf{while} $i \leq N $\\
			\hspace{3mm} Add the $i$-th task from $\bS$ to $\bDS$. \\
			\hspace{3mm} If scheduling the tasks in $\bDS$ is not viable \\
			\hspace{6mm} Remove task $i$ from $\bDS$. \\
			\hspace{3mm} $i \leftarrow i + 1$\\
			\hline
		\end{tabular}
	\end{center}
\end{table}

\section{Branch-and-Bound Method} \label{sec:BnB}

The B\&B procedure finds the optimal solution of the problem in~\eqref{eq:optimization}. This method implicitly enumerates all possible solutions on a search tree. The root node of the tree represents the whole solution space while the remaining nodes are associated with a subset of the solution space. The branching operation splits the space of the parent node into the subspaces of the resulting children nodes. Each subspace represents a partial solution. In the context of the problem in~\eqref{eq:optimization}, a partial solution is a schedule which includes a subset of the tasks. It is shown in~\cite{Jouglet11:Dominance_rules} that instead of searching for the optimal execution times of the tasks, we can search for the optimal permutation of the tasks. A schedule is obtained using the ``sequence to schedule mapping'' as explained in Section~\ref{sec:Heuristic}. In this method, each node represents a sequence of a subset of the tasks, and we search for the optimal sequence.

During the search, the nodes which provably do not contain the optimal solution are eliminated from the search tree. There are two methods for pruning nodes: dominance rules and bounds. Dominance rules are problem specific and mathematically proven rules used to eliminate nodes from the search tree. If it can be shown that the performance of all the solutions in a given node $s_1$ is worse than the performance of a solution of another node $s_2$, then $s_1$ is dominated by $s_2$, and we can therefore eliminate $s_1$ from the tree (and all its daughter nodes).

Lower and upper bounds can also be used to prune nodes in the search tree. If the lower bound on the cost of all solutions of a given node $s$ is larger than the upper bound on the cost of the optimal solution, we can conclude that $s$ does not include the optimal solution; therefore, $s$ can be eliminated from the tree. The upper bound can be initialized using a heuristic method, and it can be updated using the best-solution-found-so-far during the search.

The search tree is constructed as follows. The node at the root is an empty sequence. A branch represents choosing a task which has not been scheduled in the parent node and its deadline has not passed yet. The selected task is appended to the sequence of the parent node to obtain the sequence of the resulting child node.

A depth-first search strategy is used to traverse the nodes of the tree. When a node is pruned off, all of its children nodes are eliminated from the search. Once the entire tree has been explored, the best solution found in the search is returned.

The steps of the B\&B method are listed in Table~\ref{alg:BnB}. The lines marked with ``$*$'' are used for data recording and are not required parts of the B\&B algorithm, but will be useful for training the ML approach in the next section. The search tree is implemented using a stack data structure (see Algorithm~1 in~\cite{Jouglet11:Dominance_rules}). Here, we have modified the B\&B method of~\cite{Jouglet11:Dominance_rules} to include task dropping. Each element (node) of the stack is a tuple which consists of a sequence of tasks $\bT$ (representing a partial schedule), a set of tasks that can be scheduled right after $\bT$ (denoted by the possible-first set $\bPF$), a set of not-scheduled tasks, $\bNS$, and a set of tasks which have been dropped, $\bDR$. For each node, we maintain a set of not-dominated nodes, $\bND$, for data recording purposes.

At a given node $s$, branching is performed by choosing the task $a$ from $\bPF$ which has the earliest start time. Task $a$ is then removed from $\bPF$ and is added to $\bNS$. Note that $a$ is scheduled for the current branch and regarding the rest of the branches of $s$, it has not been scheduled yet. Therefore, it is added to the $\bNS$ set. In this way, when we add a new branch at node $s$, we merge the tasks of the $\bNS$ set with the elements of the $\bPF$ set to form the possible-first set of the new child node.

At the root node, the possible-first set is initialized to include all the tasks, and $\bNS$ and $\bDR$ are set to be empty. An upper bound $UB$ holds the cost of the best complete solution obtained during the search. $UB$ is initially set to infinity. The root node is pushed on top of the stack. Then, as long as the stack is not empty, the following procedure is performed: at each iteration, the node on top of the stack is checked. If the possible-first set $\bPF$ is empty, the node will be removed from the stack. Before removing the node, we check if there is any task in the not-scheduled set $\bNS$. If $\bNS$ is empty, the node is terminal and represents a complete solution. In this case,the overall cost $C$ is compared with $UB$. If an improvement has been achieved, the current sequence is set as the best-solution-found-so-far $\bT^*$, and $UB$ is updated accordingly. 

In the scenario that the possible-first set of the node on top of the stack is not empty, a new node is generated using a task of $\bPF$. Specifically, let $a$ be the task in $\bPF$ with the smallest start time. We remove $a$ from $\bPF$ and append it at the end of $\bT$ to obtain the sequence of the new node (denoted by $\bT'$). The possible-first set of the new node, $\bPF'$, is set as the union of $\bPF$ and $\bNS$. The dropped tasks, $\bDR'$, are inherited from $\bDR$, and the not-scheduled set of the new node, $\bNS'$, is set as empty. Then, task $a$ is added to $\bNS$. 

After a new node is generated, we determine whether it should be added to the stack for further investigation in the next iteration or it can be ignored and therefore pruned off. To do so, we first check the \textit{start-times dominance rule} on $\bT'$. This rule states that the sequence of execution times of tasks in $\bT'$ should be non-decreasing; otherwise, $\bT'$ is dominated and cannot result in an optimal solution~\cite{Yalaoui06:New_exact}. 

We next check the tasks in $\bPF'$, and any task whose deadline is before the earliest available time of the channels is dropped. At this point, the sum of the tardiness and dropping costs of the new partial solution can be computed, and it gives a lower bound on any complete solution derived from this node. If this cost is not lower than the cost of the best-solution-found-so-far, i.e., $UB$, the new node can be ignored. Furthermore, if $\bT'$ does not map to an \textit{active schedule}, it can be discarded. An active schedule is such that no task can be completed earlier without delaying another task~\cite{Jouglet04:branch}. In order to determine whether $\bT'$ is active, the tasks in $\bPF'$ are checked to see if any of them can be scheduled right before the last task in $\bT'$ (on the same timeline) without imposing a delay. 

Next, we check if $\bT'$ is a \textit{LOWS-active} (LOcally Well Sorted active) schedule~\cite{Jouglet11:Dominance_rules}. A LOWS-active schedule is such that no exchange of two adjacent tasks can improve the schedule. The adjacent tasks of task $a$ are defined as the set of tasks scheduled on the final slot of each timeline just before scheduling task $a$. The conditions for checking if a schedule is LOWS-active are the same as given in~\cite[Section~4]{Jouglet11:Dominance_rules} with the exception that we also need to consider task dropping. The conditions need to be modified such that we consider the event when a swapping of two tasks requires one of the tasks to be dropped. In this case, the availability time of the timeline is used instead of the completion time, and the dropping cost replaces the tardiness cost (since the task is not scheduled). If the new node is LOWS-active, we check if it is also a complete solution (terminal node). In this case, we update the statistics of the parent node for the purpose of data recording. This will be explained in more detail in the following section.

In the case that the bounds and dominance rules are passed, the new node is pushed on top of the stack. In this case, the current branch represents a transition to a not-dominated node. Therefore, $a$ is added to the not-dominated nodes set, $\bND$, which means that taking branch $a$ at node $s$ results in a not-dominated node. This set is only used for data recording and it is not required for the functionality of the B\&B algorithm. After this step is complete, the search goes on to the next iteration. Finally, once the entire tree is explored, the best solution found is returned. 

\begin{table}[t]
	\small
	\vspace{2.5mm}
	\caption{Branch-and-bound method}\label{alg:BnB}
	\begin{center}
		\begin{tabular}{l}
			\hline \\
			\vspace{-5mm} \\		
			\textbf{Initialization}\\
			\hspace{3mm} $UB \leftarrow \infty$ \\
			\hspace{3mm} Let $\bT$ be an empty sequence\\
			\hspace{3mm} $\bPF \leftarrow$ \{all tasks\}\\
			\hspace{3mm} $\bNS \leftarrow \{\}$ \\
			\hspace{3mm} $\bDR \leftarrow \{\}$ \\
			\hspace{3mm} $*$ $\bND \leftarrow \{\}$ \\
			\hspace{3mm} Push $(\bT, \bPF, \bNS, \bDR, \bND)$ on stack. \\
			\hline \\
			\vspace{-5mm} \\		
			\textbf{while} stack is not empty\\
			\hspace{3mm} Let $s = (\bT, \bPF, \bNS, \bDR, \bND)$ \\
            \hspace{3mm} be the node on top of stack. \\
			\hspace{3mm} \textbf{if} $\bPF \ne \{\}$ \\
			\hspace{6mm} Let $a \in \bPF$ \\
			\hspace{6mm} $\bPF \leftarrow \bPF \setminus a$ \\
			\hspace{6mm} $\bT' \leftarrow \bT | a$ \\
			\hspace{6mm} $\bPF' \leftarrow \bPF \cup \bNS$ \\
			\hspace{6mm} $\bNS' \leftarrow \{\}$ \\
			\hspace{6mm} $\bDR' \leftarrow \bDR$\\
			\hspace{6mm} $*$ $\bND' \leftarrow \{\}$\\
			\hspace{6mm} $\bNS \leftarrow \bNS \cup a$ \\
			\hspace{6mm} \textbf{if} $\bT'$ follows the start-times dominance rule \\
			\hspace{9mm} Move any task whose deadline has passed on all \\
			\hspace{9mm} timelines from $\bPF'$ to $\bDR'$. \\
			\hspace{9mm} $C' \leftarrow$ TardinessCost$\left(\bT'\right)$ + DroppingCost$\left(\bDR'\right)$ \\
			\hspace{9mm} \textbf{if} $\left(\bT' \text{ is active}\right)$ and $\left(\bT' \text{ is LOWS-active}\right)$ \\
			\hspace{12mm} $*$ \textbf{if} $\left( \bPF' = \{\} \right)$  (i.e., if the new node is terminal) \\
			\hspace{12mm} $*$ \hspace{3mm} Update the statistics of node $s$. \\
			\hspace{12mm} \textbf{if} $\left(C' < UB \right)$  \\ 
			\hspace{15mm} $*$ $\bND \leftarrow \bND \cup a$ \\
			\hspace{15mm} Push $(\bT', \bPF', \bNS', \bDR', \bND')$ on stack. \\
			\hspace{3mm} \textbf{else} \\
			\hspace{6mm} $C \leftarrow$ TardinessCost$\left(\bT\right)$ + DroppingCost$\left(\bDR\right)$ \\ 			
			\hspace{6mm} \textbf{if} $\left(\bNS = \{\} \right)$ and $\left(C < UB \right)$ \\
			\hspace{9mm} $UB \leftarrow C$ \\ 
			\hspace{9mm} $\bT^* \leftarrow \bT$ \\ 
			\hspace{6mm} $*$ \textbf{if} $s$ has termination \\
			\hspace{6mm} $*$ \hspace{3mm} Update the statistics of the parent node of $s$. \\
			\hspace{6mm} $*$ \hspace{3mm} Record the statistics of $s$. \\		
			\hspace{6mm} Remove $(\bT, \bPF, \bNS, \bDR, \bND)$ from stack. \\
			\hline
		\end{tabular}
	\end{center}
\end{table}

\subsection{Data Recording} \label{sec:DataRecord}

We record some statistics of the nodes during the search and later use them as labeled data for supervised learning of the neural networks. We keep track of the following data: for a given node $s$, we add a flag, $\mathbbmsl{I}_s$, which indicates whether we have reached at least one complete solution (terminal node) in the descendant nodes of $s$ during the search. In the case that node $s$ has termination, i.e., $\mathbbmsl{I}_s$ is true, the cost of its best terminal solution and its corresponding sequence are also recorded. 

As explained in the previous section, in the step of branching, we check whether the new child node, $s'$, is LOWS-active. If true, next we check whether $s'$ is a complete solution. If the possible-first set of $s'$ is empty, we have reached a terminal node, i.e., a complete solution. In this case, the statistics of the parent node, $s$, are updated. If $\mathbbmsl{I}_s$ is false or if it is true and the cost of $s'$ is smaller than the best-terminal-cost of $s$, $\mathbbmsl{I}_s$ will be set to true (if not already true), and the best-terminal-cost and best-terminal-sequence of $s$ will be set to the cost and sequence of $s'$, respectively.

In the B\&B procedure, before the last step of removing a given node $s$ from the stack, we check whether it has termination, i.e., if $\mathbbmsl{I}_s$ is true. In this case, the statistics of the parent node of $s$ are updated (i.e., the best-terminal-cost and best-terminal-sequence of the parent of $s$ are updated according to $s$). Furthermore, node $s$ along with all of its statistics are recorded. At this point, $\bNS$ includes all possible branches at node $s$. We also know all the tasks in $s$ which do not result in a dominated node (as given in the $\bND$ set). Furthermore, we know the best solution which can be obtained starting from node $s$ (as given in the best-terminal-sequence). Therefore, the best next task to schedule is also known. As a result, for the given node $s$, we record all possible input actions and also the optimal action to be taken next as well. In turn, the recorded labeled data can be used in supervised learning as explained in Section~\ref{sec:train}. 

\section{Monte Carlo Tree Search with Policy and Value networks} \label{sec:MCTS}

The computational burden of the B\&B method gets exponentially high as the size of the problem increases. Furthermore, the computation time of the B\&B algorithm is heavy-tailed, i.e., it can have considerable variations. For these reasons, the B\&B method is not practical for real-time applications. As we have seen, researchers have proposed approximate algorithms and heuristics as alternatives. On the other hand, as we will see, the performance of these heuristics is significantly worse than the B\&B method, motivating the search for approaches that balance performance and computational cost, especially real-time cost.

One approach to alleviate the above mentioned issues is to use the Monte Carlo tree search (MCTS) technique which can provide near-optimal solutions with reduced complexity and variation in computation time. In this section, we review the MCTS approach in its standard form; our modifications will be described in the next section. The MCTS method focuses the search on the (probabilistically) more promising nodes. Each node, as before, is associated with a subset of the solutions. In the MCTS method, each node represents a state of the system and each branch is considered as an action (decision) which causes a transition to a new state (node). In this way, finding a solution to the problem is equivalent to sequential decision making. 

The MCTS method has four steps: selection, expansion, backup, and decision making. Starting from a base node, a number of simulations are performed to analyze possible branches. This includes the first three steps of selection, expansion, and backup. These simulations create information and insight about taking different branches. After the analysis is complete, a decision is made which results in the transition to another state. The new node becomes the base for the next round and the rest of the search tree is discarded. This procedure is continued until reaching a complete solution (terminal node).

For a given complete solution $\bx$, consider a utility function $u(\bx)$ (a decreasing function of the cost of $\bx$). For a given state $s$, the \emph{value function} $v(s)$ is defined as the highest utility of the solutions that include node $s$. Furthermore, at a given node $s$, a \emph{policy} is defined as the probability distribution over taking possible actions (branches) from $s$. The optimal policy at node $s$ is the distribution that, with probability one, chooses the branch which leads to the best solution, i.e., the solution with the highest utility, $v(s)$. 

For a given edge $(s,a)$ representing branch $a$ from node $s$, the following statistics hold: a prior policy $P(s,a)$ which is the initial probability of taking action $a$, a visit count $N(s,a)$ which is the number of times branch $a$ has been selected during the previous simulations, and the action value $Q(s,a)$ which is an \emph{estimate of the expected utility} obtained by taking action $a$. 

In the selection step, the action which maximizes $Q(s,a) + U(s,a)$, where 
\[U(s,a) \propto \frac{P(s,a)}{1+N(s,a)},\] 
is taken. The first term in this selection rule, i.e., $Q(s,a)$, encourages exploitation by favoring branches with higher action values while the second term, i.e., $U(s,a)$, encourages exploration by favoring branches with lower visit counts. The algorithm keeps selecting actions until it reaches a leaf node. At this point, the selection phase of the current simulation is complete. Then, the leaf node is expanded, the statistics of its branches are initialized, and its value is estimated. 

The backup step is done by updating the statistics of the branches which have been visited during the current simulation. The visit count of these edges is incremented by one. Furthermore, the action value of each edge is updated as the average of the value of all the leaf nodes that have, so far, been encountered by taking that edge. After the backup step, the next simulation is performed. When all the simulations are done, a decision is made, which causes the transition to the next base node. A common decision rule is to take the branch with the highest visit count. Then, the next round of simulations is performed at the new base node. This procedure is repeated until reaching a complete solution, i.e., a terminal node.

Crucially, as mentioned above, in the expansion step, the value of the leaf node $s$ needs to be estimated. Furthermore, the statistics of its branches should be initialized. The visit counts and the action values of the branches of $s$ are set to zero. What remains is the initialization of the prior policy. One method to estimate the value and policy for the given node $s$ is to use \emph{policy and value networks}. It is here that a \emph{neural network} $(\bp,V) = f_{\theta}(s)$ can be used such that its input is the state $s$, and its output is the estimate of the policy and value for node $s$, respectively. The parameters $\theta$ of such a network can be trained using reinforcement learning~\cite{Silver17:AlphaGoZero}. 

We begin the training by initializing the network parameters $\theta$ randomly. For an instance of the problem, the MCTS algorithm is performed with the current policy and value network to find a solution $\bx_T$. The statistics which have been acquired during the search are now used to train the network parameters. Let $z$ be the value of the final node, i.e., $z = u(\bx_T)$. We use $z$ as the target value for the nodes which have led to the solution $\bx_T$. Furthermore, for a node $s$ on the path to $\bx_T$, define the vector $\boldsymbol{\pi}$ such that its elements are proportional to the visit counts of the branches of $s$, i.e., $N(s,a)$. Vector $\boldsymbol{\pi}$ is used as the target policy for node $s$. Then, the parameters $\theta$ are adjusted by minimizing the following overall loss 
\begin{equation}
\ell(\theta) =  (z-V)^2 - \boldsymbol{\pi}^T \log \bp + \lambda \| \theta \|^2, \label{eq:cost}
\end{equation}
where $(\bp,V) = f_{\theta}(s)$, the first and second terms of $\ell(\theta)$ are the mean-squared error and cross-entropy losses, and the last term is used for $L_2$ weight regularization controlled with parameter $\lambda$. For the next sample problem, the network with the new parameters is used. Then, the acquired data is employed to update $\theta$. This procedure is repeated to improve the performance of the network.

\section{Proposed Method} \label{sec:BnB_Policy}

Our proposed algorithm is a combination of the B\&B and MCTS methods in Sections~\ref{sec:BnB} and~\ref{sec:MCTS}. The resulting algorithm is such that we can benefit from the advantages of both methods. The bounds and dominance rules of the B\&B procedure are employed to prune off as many nodes from the search tree as possible. The MCTS method is also used to focus the search on the promising nodes. This combination will enable us to obtain near-optimal solutions while the complexity of the search is significantly reduced. 

We use the MCTS method with a number of modifications compared with the procedure explained in Section~\ref{sec:MCTS}. First, instead of stopping a simulation when reaching a leaf node, we continue the rollout until getting to a terminal node. As a result, for each simulation, the value of the final node is known (since it is a complete solution). Consequently, we do not need to use the value network for estimation (as the exact value is known). Therefore, we need only consider a policy network in our implementation of MCTS. 

Second, we keep track of different statistics to be used for the selection and decision making steps. Instead of maintaining the visit count and action value for the branches of the search tree, we hold the best-terminal-cost and best-terminal-sequence for the nodes of the tree. Furthermore, in the selection step, we only use the prior policy to take actions (visit counts and action values are not used). This last modification simplifies implementation (it is possible that better results may be achieved by considering exploration and exploitation terms in the selection step). In the decision making step, instead of taking the edge with the highest visit count, we choose the branch which leads to the best terminal node found so far during the simulations.

The steps of the proposed MCTS-based algorithm are summarized in Table~\ref{alg:MCTS}. First, the upper bound, $UB$, is set to infinity. Next, the base-node, $s$, is initialized. The sequence of tasks representing the partial schedule of $s$ is denoted by $\bT$. The set of not-dominated tasks is represented by $\bND$ which initially includes all tasks. The set of dominated tasks is denoted by $\bD$. This set will include the tasks which, during the search, will be found to not meet the bound and dominance rules. The flag $\mathbbmsl{I}_s$ indicates whether a terminal node has been reached from $s$ so far during the search. In the case that $\mathbbmsl{I}_s$ is true, $C^*$ and $\bT^*$ hold the best-terminal-cost and best-terminal-sequence obtained starting from node $s$, respectively.

The search starts from the base-node and continues for as long as the not-dominated set of the base-node is not empty. The MCTS method has two phases: simulation and decision making. In the first phase, $M$ number of rollouts are performed to gather information about the different branches of the base-node. Then, in the second phase, a branch is chosen and we move to a new base node.

For each simulation, a rollout-node, $s'$, is used to track the search along the nodes of the tree. Initially, $s'$ points to the base-node, $s$. Next, the rollout-node is expanded. Then, for as long as the not-dominated set of the rollout-node is not empty, a task is selected from it. Next, we set the rollout-node to point to the new selected node. This new node is expanded, and the selection step is repeated until reaching a terminal node. At this point, the statistics of the parent nodes of this terminal node are updated, and we move on to the next simulation. After all the simulations are complete, from the tasks in the not-dominated set of the base-node, we choose the one which leads to the best node found so far. The base-node is updated to point to the new selected node, and the procedure as explained above is repeated until the base-node reaches a complete solution. 

During the simulation phase, each visited node is expanded only once. The expansion is done by going through the tasks in the $\bND$ set (we order the tasks based on their start times). Consider a given node $s$. Each task in the $\bND$ set of $s$ represents a branch to a new node. We construct the new node and check whether it satisfies the bound and dominance rules as explained for the B\&B method. The not-dominated set of the new node, $\bND'$, is formed by merging all the elements of the $\bND$ and $\bD$ sets of $s$. 

In the case that the new node is dominated, its corresponding task in the $\bND$ set of $s$ is removed and added to the $\bD$ set of $s$. If this removal results in an empty $\bND$ set, node $s$ becomes terminal. In this case, all of the tasks in $\bD$ are dropped, the cost of the schedule associated with $s$, denoted by $C_s$, is computed and if it is lower than the upper bound, we update $UB$. Furthermore, we backup the cost and sequence of $s$ to all of its parent nodes in a way similar to how statistics of the nodes are updated for data recording as explained in Section~\ref{sec:DataRecord}. 

For each parent node of $s$, if its $\mathbbmsl{I}_s$ is false or if it is true and the best-terminal-cost of the node is larger than $C_s$, $\mathbbmsl{I}_s$ is set to true (if not already true), and the best-terminal-cost and best-terminal-sequence of that parent node is set to the cost and sequence of $s$, respectively (i.e., $C^* \leftarrow C_s$ and $\bT^* \leftarrow \bT$). The backup procedure is also performed when in the expansion step, the new created node is not dominated and it is a terminal node (its $\bND$ set is empty). In this case, the new node is indeed the best-solution-found-so-far. Therefore, the upper bound and all the parent nodes of the new node are updated accordingly.  

The maximum number of branches expanded at a given node $s$ depends on the input capacity of the neural network used to obtain the prior policy. Let $N_p$ be the input size of the network. During the expansion of $s$, at most $N_p$ number of new nodes are created. Then, the excess elements of $\bND$ are moved to the $\bD$ set. In this way, we make sure that after the expansion step is complete, the size of $\bND$ is less than or equal to $N_p$.

The last step of the expansion is to initialize the prior probabilities of the tasks in the $\bND$ set. The policy network is used for this purpose. The architecture of the policy network is explained in more detail in the next section. The input to the network is formed using the tasks in the $\bND$ and $\bD$ sets. Note that the network has a fixed input size, i.e., $N_p$. We start by taking tasks from $\bND$. If the size of $\bND$ is less than $N_p$, the rest of the tasks are taken from $\bD$, such that the overall number of tasks is less than or equal to $N_p$. The output of the network is a vector with a size equal to the number of input tasks. This vector is a probability distribution over the input tasks. From this vector, we take the elements which correspond to the tasks in $\bND$. We normalize the result to get a valid probability distribution, so that it can be used as the prior policy for the given node, $s$.

During the rollouts at a given expanded node $s$, we select the next node based on the prior policy of $s$. A task is sampled from the given prior distribution to find the next node, $s'$. In case that the upper bound is updated compared with the time that $s'$ was created and the cost of $s'$ is now larger than the upper bound, we remove the corresponding task associated with $s'$ from the $\bND$ set of $s$ and move it to the $\bD$ set of $s$. In this case, the sampling step at node $s$ is repeated.
 
\begin{table}[t]
	\small
	\vspace{2.5mm}
	\caption{Proposed MCTS method}\label{alg:MCTS}
	\begin{center}
		\begin{tabular}{l}
			\hline \\
			\vspace{-5mm} \\		
			\textbf{Initialization}\\
			\hspace{3mm} $UB \leftarrow \infty$ \\
			\hspace{3mm} $\bT \leftarrow$ empty sequence\\
			\hspace{3mm} $\bND \leftarrow$ \{all tasks\}\\
			\hspace{3mm} $\bD \leftarrow \{\}$ \\
			\hspace{3mm} $\mathbbmsl{I}_s \leftarrow$ false \\
			\hspace{3mm} $C^* \leftarrow \infty$ \\
			\hspace{3mm} $\bT^* \leftarrow$ empty sequence\\
			\hspace{3mm} base-node: $s$ $\leftarrow$ $(\bT, \bND, \bD, \mathbbmsl{I}_s, C^*, \bT^*)$\\
			\hline \\
			\vspace{-5mm} \\		
			\textbf{while}  not-dominated set of base-node is not empty\\
			\hspace{3mm} \textbf{for} $M$ number of simulations \\
			\hspace{6mm} Set rollout-node to point to the base-node: $s' \leftarrow s$  \\
			\hspace{6mm} Expand the rollout-node.\\
			\hspace{6mm} \textbf{while}  not-dominated set of rollout-node is not empty\\
			\hspace{9mm} Select a branch from not-dominated set of rollout-node.\\
			\hspace{9mm} Set rollout-node to point to the new selected node.\\
			\hspace{9mm} Expand the new rollout-node.\\
			
			\hspace{3mm} Decision making: \\
			\hspace{6mm} Using $\bT^*$ of the base-node, take the branch which leads\\
			\hspace{6mm} to the best node found during the search. \\			
			\hspace{3mm} Set base-node to point to the new selected node.\\
			\hline
		\end{tabular}
	\end{center}
\end{table}

\subsection{Policy Network Architecture}

As mentioned in the previous section, in the expansion step of the MCTS algorithm, we use the policy network to provide the prior probability distribution over the tasks in the not-dominated set, $\bND$, of the given node, $s$. The input to the network is the state of the node, and the output is the prior policy, i.e., $\bp = f_{\theta}(s)$, where $\theta$ denotes the weights of the network.

The input to the network is treated as an image with height $1$, width $N_p$, and $N_c$ channels (features). Each column of the input image is associated with a task, and for each task, we have the $N_c$ features as given in Table~\ref{tbl:features}. Here, $N_c = 8 + K$ where $K$ is the number of available timelines (See Table~\ref{tbl:features} for a list of the 8 features.). 

The input tasks are taken from the not-dominated, $\bND$, and dominated, $\bD$, sets, and the scheduled or dropped tasks are not used. At a given node, the overall number of tasks in $\bND$ and $\bD$ may be smaller, equal, or greater than $N_p$. However, the number of input tasks to the network is fixed to $N_p$. To deal with this issue, the first three features are used to indicate whether a given input task is active or it should be ignored. If the task is taken from $\bD$, the first three features are set to ``$1, 0, 0$''. If the task is taken from $\bND$, the first three features are set to ``$0, 1, 0$''. Otherwise, the first three features are set to ``$0, 0, 1$'' to indicate that this input task should be ignored. Note that, in a convolutional neural network, we cannot use one indicator variable for the 3 possibilities since linear combinations of any two possibilities should not lead to the third. For each input task, the features $4$ to $8$ are taken from the parameters of that task. If the input is not active, these features are set to zero. Finally, the time availability indicators of the timelines at the given input node, are set as the last four features. Note that these values are the same for all the input tasks. 

To form the input of the network, we start with tasks from $\bND$. If the size of $\bND$ is less than $N_p$, the rest of the tasks are taken from $\bD$, such that the overall number of tasks is less than or equal to $N_p$. After, the tasks are selected, they are sorted based on their start times and then put together to form the input image. Then, input features $4$ to $12$ are normalized by a fixed number which represents the largest possible value for these parameters. Therefore, after normalization, all the input features are less than or equal to $1$. 

As mentioned in the previous section, the output of the network is a vector of length equal to the number of active tasks. From this vector, we take the elements corresponding to the not-dominated input tasks. We normalize the result to get a valid probability distribution for use as the prior policy for the given node $s$.

\begin{table*}[t]
	\small	
	\vspace{2.5mm}
	\caption{Features of the tasks as used for the input of the policy network} \label{tbl:features}
	\vspace{-1mm}
	\begin{center}
		\begin{tabular}{l l}
			Feature & Description \\
			\hline 
			$1, 2, 3$ & status ($1,0,0$: dominated, $0,1,0$: not-dominated, $0,0,1$: ignore this input task) \\
			$4$ & start time, $r_n$ \\
			$5$ & deadline, $d_n$ \\
			$6$ & length, $\ell_n$ \\
			$7$ & tardiness weight, $w_n$ \\
			$8$ & dropping cost, $D_n$ \\
			$9, 10, 11, 12$  &  timeline availability indicators, $g_1, g_2, g_3, g_4$ \\
			\hline 
		\end{tabular}
	\end{center}
\end{table*}

We implement the policy network using a convolutional neural network. The motive for using this type of network is the translation invariance property which helps recognize patterns between neighboring tasks \emph{regardless of their position in the input image}. Here, we implement the policy network with $7$ layers. The first four layers are convolutional and the last three layers are fully connected. 

\emph{Convolution}: Each convolutional layer is composed of the following operations. First, the input is convolved with a filter. Next, we perform batch normalization, which will be explained in more detail in the sequel. Then, the result goes through a non-linear rectifier function (ReLU). In the training stage, we also perform the dropout operation before passing the result to the next layer. 

At all the convolutional layers the kernel size is $1 \times 7$, i.e., the filter has a height of $1$ and width of $7$. Therefore, we are looking at the features of $7$ consecutive tasks at each stride. The first filter has an input depth of $N_c$ (which is equal to $12$ in our implementation) and an output depth of $96$. The rest of the filters have the same input and output depth of $96$. We use valid padding for the convolution operations, i.e., the filters do not go past the edges of the input, as there is no zero padding. Furthermore, the convolutions are performed with a stride of $1$.

The output of the fourth layer is vectorized and then passed to a fully connected layer with $2048$ hidden units. The second fully connected layer has $1024$ hidden units, and the output of the last layer has a length of $N_p$. The first two fully connected layers have the following components. First, the input is multiplied by a weight matrix (note that there is no bias term at this stage). Next, we perform batch normalization. Then, the result goes through the ReLU activation function. In the training phase, we also perform the dropout operation before passing the result to the next layer. The last fully connected layer simply is a linear transformation with a weight matrix and a bias term. 

The result of the last layer has a length of $N_p$. We truncate this output such that it has a length equal to the number of active inputs (input tasks whose third feature is zero; see Table~\ref{tbl:features}). Finally, the result goes through the softmax function to produce the output of the network. 

\emph{Dropout}: During the training phase, the dropout operation is performed for the purpose of regularization. This is done to avoid the problem of overfitting the neural network to the training samples. Regularization methods help the network to learn the general patterns in the training samples instead of memorizing them. The dropout technique is one of the most commonly used methods for regularization~\cite{Hinton14:Dropout}. During the training stage, a fixed percentage (50 percent in our case) of the units of each layer is randomly dropped (multiplied by zero). This results in training subnetworks of the base network. Dropout regularizes each unit to be not merely a good feature but a feature that is good in many contexts.

Batch normalization is also used to improve the optimization process during the training phase~\cite{Ioffe15:batch_normalization}. Let $\bH$ be a mini-batch of activations of a given layer. To normalize $\bH$, we replace it with 
\begin{equation}
\bH' = \frac{\bH - \boldsymbol{\mu}}{\boldsymbol{\sigma}} \label{eq:batch_norm}
\end{equation}
where $\boldsymbol{\mu}$ and $\boldsymbol{\sigma}$ represent the mean and standard deviation of the units, respectively. For the convolutional layers, these moments are obtained over the first three dimensions of $\bH$, and therefore, they have a length of $96$ (the depth of the feature maps). For the fully connected layers, the moments are computed over the rows of $\bH$, and therefore, their length is equal to the number of hidden units. The subtraction and division operations in~\eqref{eq:batch_norm} are performed by broadcasting. 

During the training, $\boldsymbol{\mu}$ and $\boldsymbol{\sigma}$ are obtained using the sample mean and variance of the current batch~\cite[Section~8.7.1]{Goodfellow-et-al-2016}. During the inference, $\boldsymbol{\mu}$ and $\boldsymbol{\sigma}$ can be replaced by the running averages that were collected during training time.

Normalizing the activations of the layers may reduce the expressive power of the network~\cite[Section~8.7.1]{Goodfellow-et-al-2016}. Therefore, the normalized batch $\bH'$ is replaced with $\boldsymbol{\gamma} \bH' + \boldsymbol{\beta}$ where $\boldsymbol{\gamma}$ and $\boldsymbol{\beta}$ are variables learned during the training phase. 

\subsection{Training}\label{sec:train}

Training data is obtained using the B\&B method as explained in Section~\ref{sec:DataRecord}. For a given sample problem, all the nodes resulting in a complete solution (terminal node) are recorded. The data for each node includes all the possible tasks which can be scheduled next after the given node. These tasks are listed in the $\bNS$ set. Furthermore, all the tasks which are not dominated are given in the $\bND$ set. Then, the set of dominated tasks, $\bD$, is obtained as the difference between the $\bNS$ and $\bND$ sets. 

The recorded data of a node also includes its best-terminal-sequence. From this sequence, we can determine the best next task, $a^*$, to be scheduled after the sequence of the given node. Therefore, for each node $s$, we have the state-action pair, $(s,a^*)$, which can be used as labeled data for supervised learning of the policy network. For the purpose of training the network, $a^*$ is represented by one-hot encoding in a vector, $\bp^*$, which has the same length of the network output. In this representation, the element corresponding to task $a^*$ is set to one, and the rest of the elements are set to zero.

In order to generate the training data, we run the B\&B method for $1000$ sample problems with $N=40$ tasks and $K=4$ channels (timelines). We consider a time window of $100$~ms. The start time of the tasks is uniformly distributed on the time window, i.e., $r_n \sim \mathcal{U}(0,100)$, where $x \sim \mathcal{U}(a,b)$ represents a random variable uniformly distributed between $a$ and $b$. For each task, the interval between the start time and the deadline, i.e., $d_n - r_n$, is sampled from $\mathcal{U}(2,12)$. The task length, $\ell_n$, is distributed according to $\mathcal{U}(2,11)$. Furthermore, the dropping costs, $D_n$, and the tardiness weights, $w_n$, have distributions $\mathcal{U}(100,500)$ and $\mathcal{U}(1,5)$, respectively.

For each instance of the problem, let the number of recorded nodes be denoted by $N_s$. We sample $1000$ nodes from these $N_s$ nodes, and if $N_s < 1000$, all of the nodes are taken. Next, the sampled nodes from all of the instance problems are randomly permuted and then put together to form a dataset which ends up to have a size of $530976$. We have used $500,000$ and $25,000$ samples for the training and validation datasets, respectively. The rest of the samples are used for testing the trained network.

The parameters of the policy network, $\theta$, are trained by minimizing the average cross-entropy loss given by
\begin{equation}
\ell_b = -\frac{1}{N_b} \sum_{i = 1}^{N_b} \boldsymbol{\bp_i^*}^T \log \bp_i \label{eq:avg_loss}
\end{equation}
where $N_b = 64$ is the mini-batch size, $ \left( s_i, \bp_i^* \right)$ is the $i$-th row of the training mini-batch, and $\bp_i=f_{\theta}(s_i)$. The input features to the network, i.e., the task parameters and the timeline availability indicators, are normalized by the maximum possible value of the parameters which is $500$.  

We minimize the loss given in~\eqref{eq:avg_loss} using the Adaptive Moment Estimation (Adam) optimization method~\cite{Jimmy14:Adam}. We train the network using $800000$ steps of the Random Reshuffling (RR) technique~\cite{Ozdaglar15:RR} with a learning rate of $0.001$. 

\begin{figure}[t]
	\centering{\includegraphics[width=15cm]{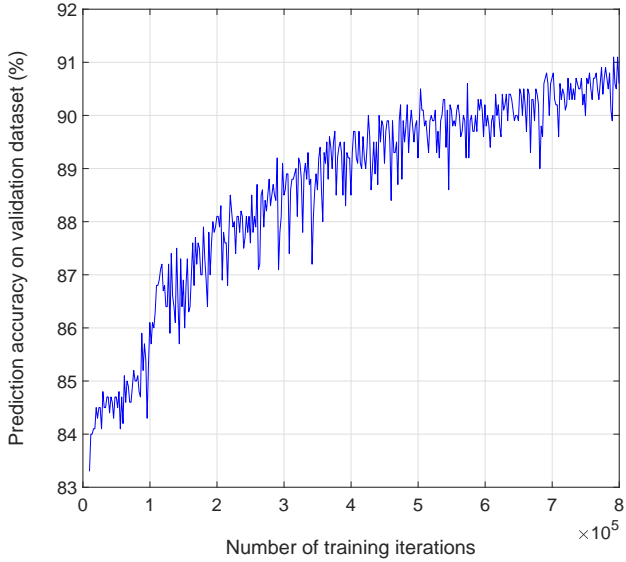}}
	\caption{Prediction accuracy on validation dataset versus the number of training iterations. Each iteration is done using a mini-batch of size $64$ on the training dataset.} \label{fig:valid_accuracy}
\end{figure}

The performance of the policy network on the validation dataset during the training is illustrated in Fig.~\ref{fig:valid_accuracy}. For a given pair from the validation dataset, $ \left( s, \bp^* \right)$,  a correct prediction represents the case when  $\arg\max \bp = \arg\max \bp^*$, where $\bp=f_{\theta_t}(s)$ and $\theta_t$ denotes the parameters of the network at the $t$-th iteration. A prediction accuracy of about $91$\% is achieved after the $800000$ iterations. 

Before ending this section, it is worth commenting on the differences between the proposed approach and that in our previous work~\cite{MSH18:RadarConf}. As mentioned in the introduction, this previous work is only valid for a fixed number of tasks, whereas the method proposed here can be applied to an arbitrary number of tasks. In~\cite{MSH18:RadarConf} we use a \emph{value network} (a deterministic approach) to prune nodes, whereas, here, the policy network helps select actions at a node. 

Importantly, in the previous approach, the entire tree must be traversed (with B\&B pruning), thereby limiting the computational gains over B\&B. Here, after a decision is made at a base node, the rest of the tree is discarded. However, the fundamental difference is that the ML approach in~\cite{MSH18:RadarConf} determines an action at each node, whereas here, the MCTS provides a \emph{probability distribution}.

\section{Performance Evaluation}\label{sec:sim}

Having described how the policy network is used in conjunction with the B\&B method and how it is trained, we now illustrate the performance of the proposed method. In this regard, the performance of the heuristics and the B\&B algorithm act as baselines. We consider a multichannel radar with $K = 4$ identical channels. There are different numbers of tasks distributed over a timeline window of $100$~ms. The objective is to schedule the tasks such that the overall cost of dropping and delaying the tasks is minimized (see~\eqref{eq:optimization}). 

The distributions of the task parameters are the same as the ones given for the training data of the policy network as explained in Section~\ref{sec:train}. For each task, the start time $r_n$, the interval between the start time and the deadline, i.e. $d_n - s_n$, the task length $\ell_n$, the dropping cost $D_n$, and the tardiness weight $w_n$ are distributed according to $\mathcal{U}(0,100)$, $\mathcal{U}(2,12)$, $\mathcal{U}(2,11)$, $\mathcal{U}(100,500)$, and $\mathcal{U}(1,5)$, respectively. We perform the simulations for $1000$ sample problems to obtain the average performance of the considered methods. For each simulation scenario, the same set of parameters is used for all the algorithms.

\begin{figure}[t]
	\centering{\includegraphics[width=15cm]{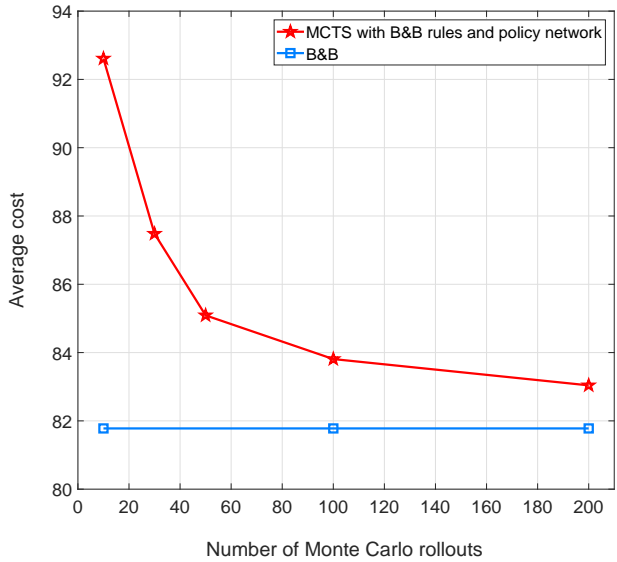}}
	\caption{Average cost versus the number of Monte Carlo rollouts, $M$. The number of tasks is $N=40$.} \label{fig:cost_mc}
\end{figure}

\emph{Rollouts:} In the first experiment, we show how the number of Monte Carlo rollouts, $M$, affects the performance and complexity of the MCTS method as proposed in Section~\ref{sec:BnB_Policy}. The average cost versus $M$ is illustrated in Fig.~\ref{fig:cost_mc}. The number of tasks is fixed to $N=40$. The curve marked with stars represents the proposed method. The performance of the B\&B algorithm, which is optimal, is also plotted for comparison. It can be seen that with as low as $M=10$ Monte Carlo rollouts, the performance of the proposed MCTS algorithm is close to optimal, and it further improves by increasing $M$. From Fig.~\ref{fig:cost_mc}, we can observe that with about $M=50$, we can get a near-optimal result. 

\emph{Average Cost:} The average cost of the heuristic methods, as explained in Section~\ref{sec:Heuristic}, are given in Table~\ref{tbl:cost_n40_mc50}. The number of tasks is fixed to $N=40$. For the MCTS method, we provide results with $M=50$ rollouts. It can be seen that the gap between the performance of the heuristic methods and that of the B\&B method is quite large, while the propose MCTS algorithm has a near-optimal average cost. Using the task swapping (SW) technique improves the performance of the heuristic methods to some extent. However, the gap still remains quite significant. This table is the key result of this paper.

\begin{figure}[t]
	\centering{\includegraphics[width=15cm]{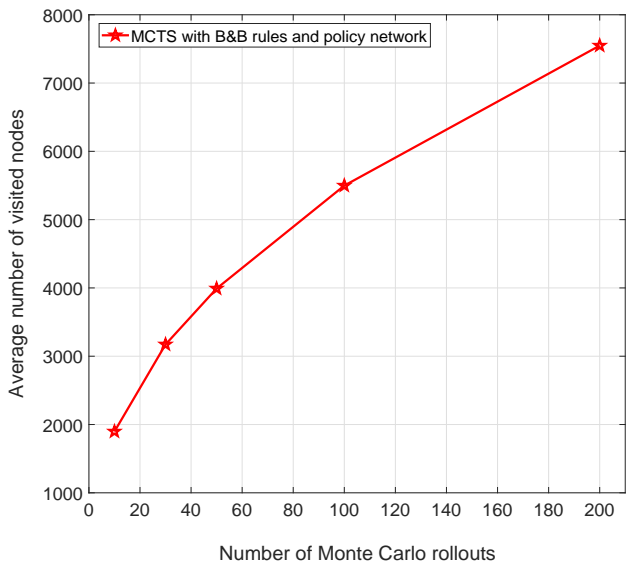}}
	\caption{Average number of visited nodes for the MCTS method versus the number of Monte Carlo rollouts, $M$. The number of tasks is $N=40$.} \label{fig:num_nodes_mc}
\end{figure}

\emph{Complexity:} Next, we analyze the complexity of the proposed MCTS method. The average number of nodes which have been visited during the search versus various number of Monte Carlo rollouts is depicted in Fig.~\ref{fig:num_nodes_mc}. The number of tasks is again fixed to $N=40$. For comparison, the average number of visited nodes by the B\&B algorithm is $669738$, which is about $2$ \emph{orders of magnitude} larger than the visited nodes by the MCTS method. If using sequential processing, wherein complexity is essentially linear in the number of nodes visited, this implies two orders of magnitude savings in execution complexity. 

\begin{figure}[t]
	\centering{\includegraphics[width=15cm]{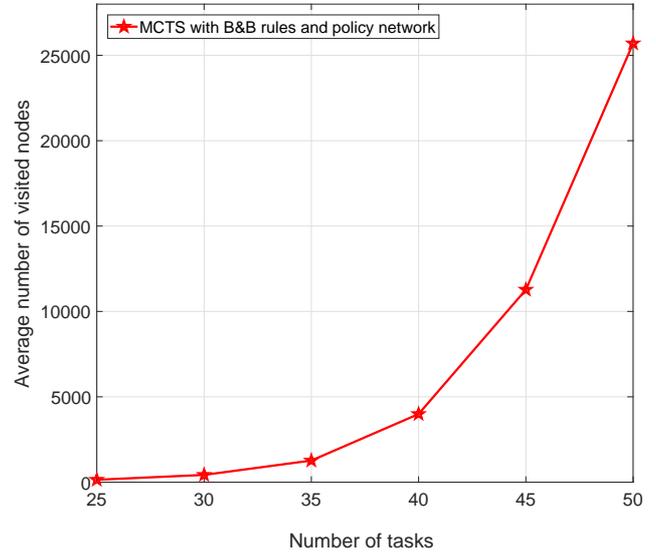}}
	\caption{Average number of visited nodes for the MCTS method versus the number of tasks, $N$. The number of Monte Carlo rollouts is $M=50$.} \label{fig:num_nodes_n}
\end{figure}

The average number of visited nodes versus the number of tasks for the MCTS method is plotted in Fig.~\ref{fig:num_nodes_n}. The number of Monte Carlo rollouts is fixed to $M=50$. For comparison, the average number of visited nodes by the B\&B algorithm is given in Table~\ref{tbl:num_nodes_n25_50_mc50}. It can be seen that for $N=45$ tasks, the complexity of the B\&B algorithm is more than $3$ orders of magnitude larger than that of the proposed MCTS method.

\begin{figure}[t]
	\centering{\includegraphics[width=15cm]{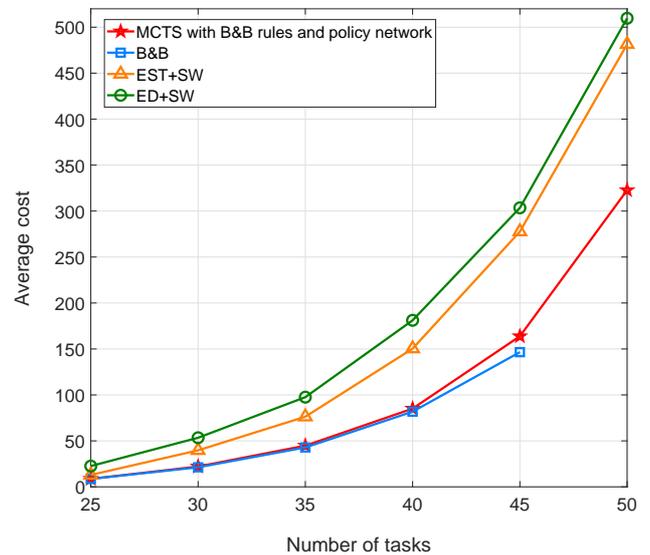}}
	\caption{Average cost versus the number of tasks, $N$. For the MCTS method, number of Monte Carlo rollouts is set to $M=50$. SW represents task swapping.} \label{fig:cost_n}
\end{figure}

\emph{Varying Number of Tasks:} In the second experiment, we investigate the performance of the proposed method for different number of tasks, $N$. The number of Monte Carlo rollouts is fixed to $M=50$ in this scenario. The average cost of the MCTS, B\&B, and the heuristic methods versus the number of tasks is illustrated in Fig.~\ref{fig:cost_n}. The performance of the B\&B method is obtained up to $N=45$ tasks. Beyond this point, the complexity of the B\&B algorithm grows extremely large. As can be seen in Fig.~\ref{fig:cost_n}, the performance of the proposed MCTS method is very close to the optimal method. Note that the policy network used in the MCTS algorithm is trained on problems with $N=40$ tasks. However, the proposed algorithm is designed such that it can handle problems with larger number of tasks as well. It can be seen that for $N=45$, the proposed method still has near-optimal performance.

\begin{figure}[t]
	\centering{\includegraphics[width=15cm]{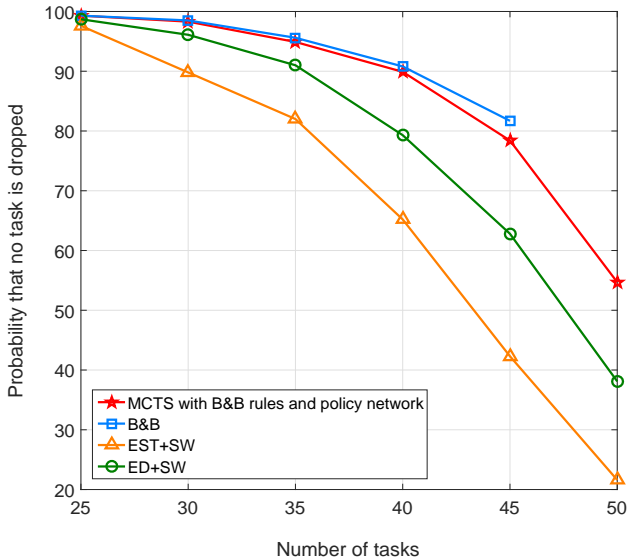}}
	\caption{Probability that all the tasks are scheduled with any dropping versus the number of tasks, $N$. For the MCTS method, number of Monte Carlo rollouts is set to $M=50$. SW represents task swapping.} \label{fig:PViable_n}
\end{figure}

Finally, we compare the MCTS, B\&B, and the heuristic methods with respect to their ability of scheduling all the tasks without dropping any of them. The probability that no task is dropped versus the number of tasks is depicted in Fig.~\ref{fig:PViable_n}. As it can be seen, the performance of the proposed MCTS method is very close to the B\&B algorithm and significantly better than the heuristic methods.

\begin{table*}[t]
	\small	
	\vspace{2.5mm}
	\caption{Average cost of various methods. The number of Monte Carlo rollouts is set to $M=50$, and the number of tasks is $N=40$. SW represents task swapping.} \label{tbl:cost_n40_mc50}
	\vspace{-1mm}
	\begin{center}
		\begin{tabular}{l l l l l l}
			MCTS & B\&B & EST & EST+SW & ED & ED+SW \\
			\hline 
			$85.1$  & $81.8$ & $198.1$ & $150.4$ & $198.2$ & $181.2$ \\
			\hline 
		\end{tabular}
	\end{center}
\end{table*}

\begin{table*}[t]
	\small	
	\vspace{2.5mm}
	\caption{Average number of visited nodes versus the number of tasks, $N$. The number of Monte Carlo rollouts is set to $M=50$.} \label{tbl:num_nodes_n25_50_mc50}
	\vspace{-1mm}
	\begin{center}
		\begin{tabular}{l l l l l l l}
			 & $25$ & $30$ & $35$ & $40$ & $45$ & $50$ \\
			\hline 
			MCTS & $140$ & $425$ & $1266$ & $3991$ & $11274$ & $25699$ \\
			B\&B & $173$ & $1031$ & $27068$ & $669738$ & $13622348$ \\
			\hline 
		\end{tabular}
	\end{center}
\end{table*}

\section{Conclusion}\label{sec:conclude}

In this paper, we consider the problem of radar resource management for a multichannel multifunction radar. We introduce how machine learning techniques can be used to acquire knowledge during the operation of the cognitive radar in a way that such knowledge can be used in future decision making. 

The branch-and-bound (B\&B) algorithm finds the optimal solution for the task scheduling problem but with high computational burden. We show how the solutions obtained from the offline execution of the B\&B method can be used to train neural networks which can help to reduce the complexity of the search. We propose a method based on the Monte Carlo tree search which along with dominance rules of the B\&B method uses a policy network to focus the search on more promising branches. The proposed method has near-optimal performance, while the computational complexity is significantly lower than the B\&B method.

It is worth emphasizing that our performance analysis assumes that the probability distributions of task parameters in the training and inference phases are the same. The robustness of the approach to possible differences between the two phases is unclear. If the distributions of task parameters change over time, the approach would have to change to use the performance of recent executions to continuously train the neural network. However, the overall architecture of our solution framework should not change.

\vspace{-1mm}

\section{Acknowledgments}\label{sec:ack}

This work is supported by the Natural Sciences and Engineering Research Council (NSERC) of Canada and Defence Research and Development Canada (DRDC).

\bibliographystyle{IEEEtran}

\vfill\pagebreak

\end{document}